\documentclass[runningheads]{llncs}
\usepackage[T1]{fontenc}
\usepackage{graphicx}
\usepackage[colorlinks=true,linkcolor=blue, citecolor=blue, urlcolor=blue]{hyperref}
\usepackage{multirow}

\begin{document}
\title{Interactive Segmentation Model for Placenta Segmentation from 3D 
 Ultrasound images} 
\titlerunning{Interactive Placenta Segmentation from 3D Ultrasound images}

\author{Hao Li\inst{1} \and
Baris Oguz\inst{2} \and
Gabriel Arenas\inst{2} \and
Xing Yao\inst{1} \and
Jiacheng Wang\inst{1} \and
Alison Pouch\inst{2} \and
Brett Byram\inst{1} \and
Nadav Schwartz\inst{2} \and
Ipek Oguz\inst{1}}

\authorrunning{H. Li et al.}

\institute{Vanderbilt University \and
University of Pennsylvania}


%
\maketitle              
\begin{abstract}

Placenta volume measurement from 3D ultrasound images is critical for predicting pregnancy outcomes, and manual annotation is the gold standard. However, such manual annotation is expensive and time consuming. Automated segmentation algorithms can often successfully segment the placenta, but these methods may not consistently produce robust segmentations suitable for practical use. Recently, inspired by the Segment Anything Model (SAM), deep learning-based interactive segmentation models have been widely applied in the medical imaging domain. These models produce a  segmentation from visual prompts provided to indicate the target region, which may offer a feasible solution for practical use. However, none of these models are specifically designed for interactively segmenting 3D ultrasound images, which remain challenging due to the inherent noise of this modality. In this paper, we evaluate publicly available state-of-the-art 3D interactive segmentation models in contrast to a human-in-the-loop approach for the placenta segmentation task. The Dice score, normalized surface Dice, averaged symmetric surface distance, and 95-percent Hausdorff distance are used as evaluation metrics. We consider a Dice score of 0.95 a successful segmentation. Our results indicate that the human-in-the-loop segmentation model reaches this standard. Moreover, we assess the efficiency of the human-in-the-loop model as a function of the amount of prompts. Our results demonstrate that the human-in-the-loop model is both effective and efficient for interactive placenta segmentation. The code is available at \url{https://github.com/MedICL-VU/PRISM-placenta}.

\end{abstract}

\section{Introduction}
Placenta volume measurement from 3D ultrasound (3DUS) images is associated with  fetal size \cite{schwartz2012two} 
and adverse pregnancy outcomes, such as preeclampsia \cite{redman2005latest,schwartz2014first} and intrauterine growth restriction \cite{biswas2008gross}, which contribute  to perinatal morbidity and mortality.
Manual annotations of the placenta are considered the gold standard in practice to ensure precise measurements. However, obtaining annotations is time-consuming, subjective, and requires expert knowledge. In contrast, fully automated algorithms using multi-atlas segmentation techniques have been developed to segment the placenta \cite{oguz2016automated}. In more recent years, deep learning-based automated approaches have demonstrated state-of-the-art performance as the leading method for placenta segmentation \cite{looney2018fully,looney2021fully,looney2017automatic,oguz2018combining,pouch2020automated,schwartz2022fully,zimmer2023placenta}. Despite their advancements, these automated segmentation methods may not consistently deliver high performance in every case, especially in poor quality images with high levels of noise or artifacts. Robust segmentation of the placenta remains challenging due to unclear boundaries in 3DUS images \cite{wang2021deep}, especially during the early pregnancy when the placenta and uterine tissue are less distinct. Additionally, uterine contractions can cause significant  anatomical variations \cite{fiorentino2023review}. This issue is further complicated by the weak contrast and inherent noise in 3DUS images, as well as shadow  \cite{xu2021shadow} and attenuation artifacts common in posterior placentas.

An interactive segmentation method could serve as a potential alternative for addressing these variabilities and improving the robustness of placenta segmentation. Practical uses of such a method would include facilitating annotation of large datasets for training automated models, as well as segmenting challenging images where automated methods may fail. Early work in this direction includes the VOCAL platform from GE-Healthcare, which involves interpolation between manually labeled slices, a random-walker algorithm \cite{STEVENSON20153182}, as well as a manually initialized multi-atlas label fusion method   \cite{oguz2020minimally}. More recently, the deep learning-based interactive segmentation method, Segment Anything Model (SAM) \cite{Kirillov_2023_ICCV}, has shown generalizability and ability to produce precise segmentation and has been widely adopted in the medical domain \cite{ma2024segment,li2023promise,gong20233dsam,wang2023sam,cheng2023sam,li2024prism}, including for US applications \cite{yao2023false,lin2023samus,tu2024ultrasound}. These models either adapt pretrained weights from SAM or they are trained from scratch for the medical imaging domain. These SAM-based models require users to provide prompts, such as points \cite{li2023promise,gong20233dsam,wang2023sam,cheng2023sam,li2024prism,tu2024ultrasound,lin2023samus,wang2023novel}, boxes \cite{yao2023false,ma2024segment,li2024prism}, scribbles \cite{wong2023scribbleprompt,li2024prism}, and masks \cite{wang2023sam,cheng2023sam,li2024prism} for interactions.  


However, a robust interactive segmentation model should be both \underline{effective} \textit{and} \underline{efficient} 
in responding to user prompts with minimal interactions. Unfortunately, even with such interactions, these current studies designed for US applications may not always provide consistent, robust segmentation for practical use. Moreover, 2D interactive models might not be as efficient for 3D images, which may require visual prompts within each slice. Even if prompts can be propagated accurately between slices to reduce user burden, a 2D model also cannot capture depth information as effectively as 3D models, which may limit the segmentation performance. Currently, there is no 3D interactive model designed for placenta segmentation that meets these goals for effectiveness and efficiency.

The current publicly available 3D interactive segmentation models aim to achieve robust performance by adapting pretrained weights from SAM \cite{li2023promise,gong20233dsam} and incorporating a human-in-the-loop design for iterative corrections \cite{wang2023sam,li2024prism}. Among these methods, PRISM \cite{li2024prism} stands out as it achieves human-level performance, making it an effective interactive segmentation model suitable for practical applications. Unlike other models, PRISM supports a wide range of prompt types, allowing for precise segmentation adjustments until user expectations are met. However, PRISM was not originally designed for placenta segmentation or for 3DUS data, and its effectiveness in this application remains untested.

In this paper, we evaluate the publicly available state-of-the-art 3D interactive segmentation models \cite{li2023promise,gong20233dsam,wang2023sam}, in contrast to the human-in-the-loop PRISM algorithm \cite{li2024prism} for placenta segmentation from 3DUS images. We focus on both effectiveness and efficiency, as our goal is to identify a model suitable for practical use. We consider a Dice score of 0.95 as a  bar for success, which is conservative given previous reports of interrater variability in the 0.85-0.90 range for manual segmentations \cite{zimmer2023placenta}. Our comprehensive experiments show PRISM \cite{li2024prism} has superior performance and reaches the success bar within only a few iterations. 
Our findings confirm that PRISM is effective and efficient, making it suitable for practical use.

\section{Materials and Methods}
\subsection{Datasets}
3D ultrasound volume datasets (n=124) were acquired from women at 11–14 weeks of gestation using GE Voluson E8 ultrasound machines. The dimensions of the raw images ranged from $245 \times 265 \times 173$ to $714 \times 726 \times 488$ voxels, with a mean isotropic resolution of $0.49 \pm 0.04 \mathrm{mm}$. The training set consisted of 100 subjects with 25 used for validation, and the remaining 24 were used for testing (14 anterior/10 posterior). We resampled all images to a $1$mm isotropic resolution, performed intensity clipping based on the $0.5^{th}$ and $99.5^{th}$ percentiles of the foreground, and applied Z-score normalization based on foreground voxels. Random zoom and intensity shift were used as data augmentations.

\begin{figure}[t]
\centering
\includegraphics[width=\linewidth]{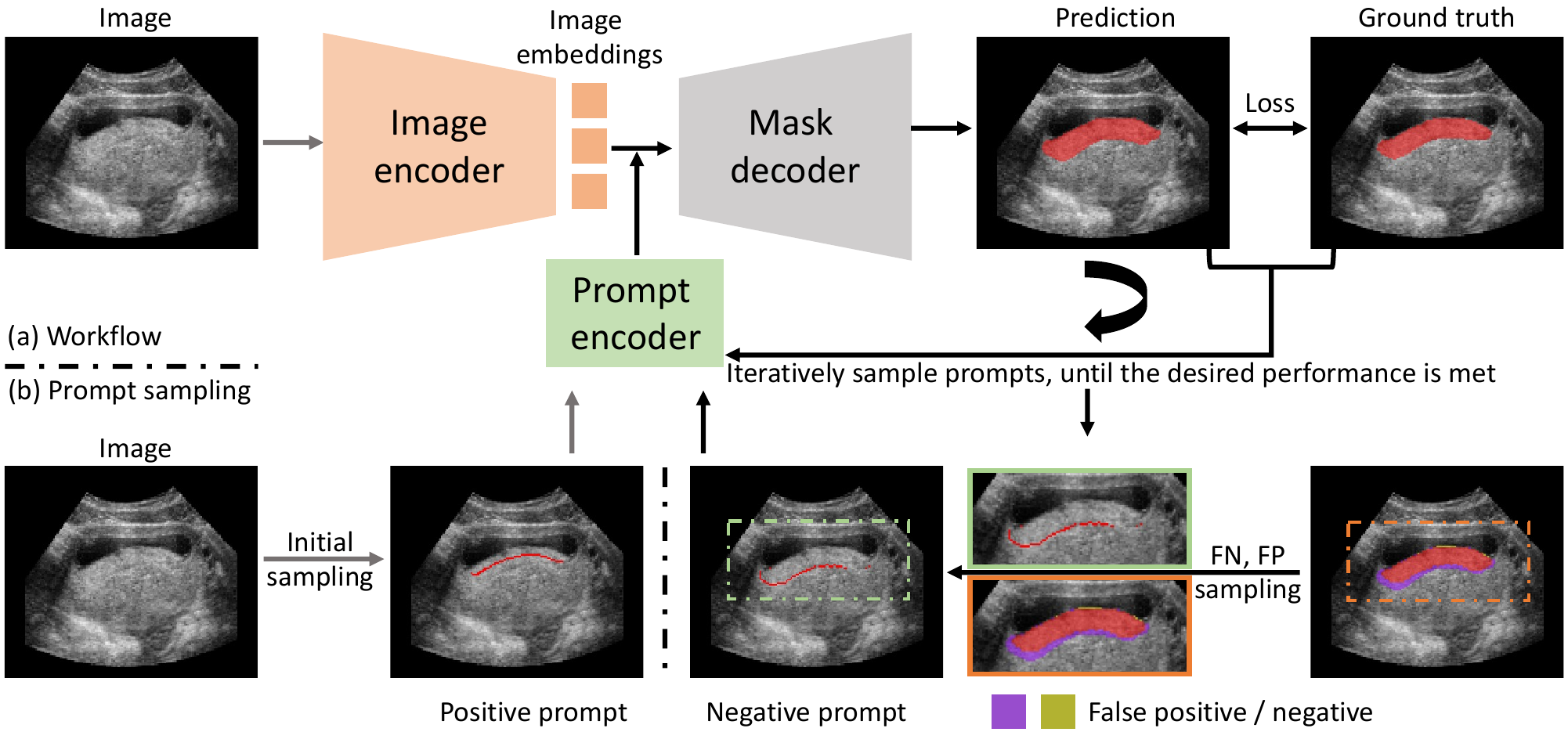}   
\caption{(a) The 3D interactive segmentation model (PRISM \cite{li2024prism}), illustrated in 2D. (b) Prompt sampling for positive (left) and negative prompts (right). To mimic human behavior, we sample prompts from the FN and FP regions of the current segmentation at each iteration. The initial sampling only has positive prompts. }
\label{framework}
\end{figure}

\subsection{Interactive segmentation model}
We adopt PRISM \cite{li2024prism}, a human-in-the-loop approach 3D interactive segmentation model (Fig.~\ref{framework}(a)), for the 3DUS placenta segmentation task.  For each image, along with visual prompts, image and prompt encoders are used to produce respective embeddings. Specifically, the image encoder is a hybrid model that combines parallel convolutional and transformer paths \cite{li2022cats}, and the prompt encoder contains several embedding layers. These embeddings interact through self- and cross-attention mechanisms before being fed into a CNN-based decoder to generate the segmentation.


As a human-in-the-loop design, PRISM employs iterative learning to achieve improvements through successive iterations. To mimic human behavior in our experiments, the positive and negative visual prompts are generated from false negatives (FN) and false positives (FP) of the segmentation result at the previous iteration, respectively. For the initial sampling (Fig.~\ref{framework}(b), left), only false negatives are considered, and the positive prompts are used to indicate the target region. We use a constant number of iterations (n=11) for all subjects for training and inference in our experiments. However, in practice, this number would vary between subjects, as the user would stop when the segmentation result meets their expectation. 



\subsection{Visual prompt generation}
\label{prompt_generation}
PRISM can take various visual prompts as additional input along the input image, including points, boxes, and scribbles. We generate these different visual  prompts to indicate the target region as follows:

\noindent \underline{\textbf{Point}}: At both initial and subsequent iterations, point prompts are randomly sampled with uniform distribution from the FN and FP regions. 

\noindent    \underline{\textbf{Box}}: The 3D bounding box is determined based on the ground truth and represented as two points. It is only sampled at initial iteration.

\begin{figure}[t]
\centering
\includegraphics[width=\linewidth]{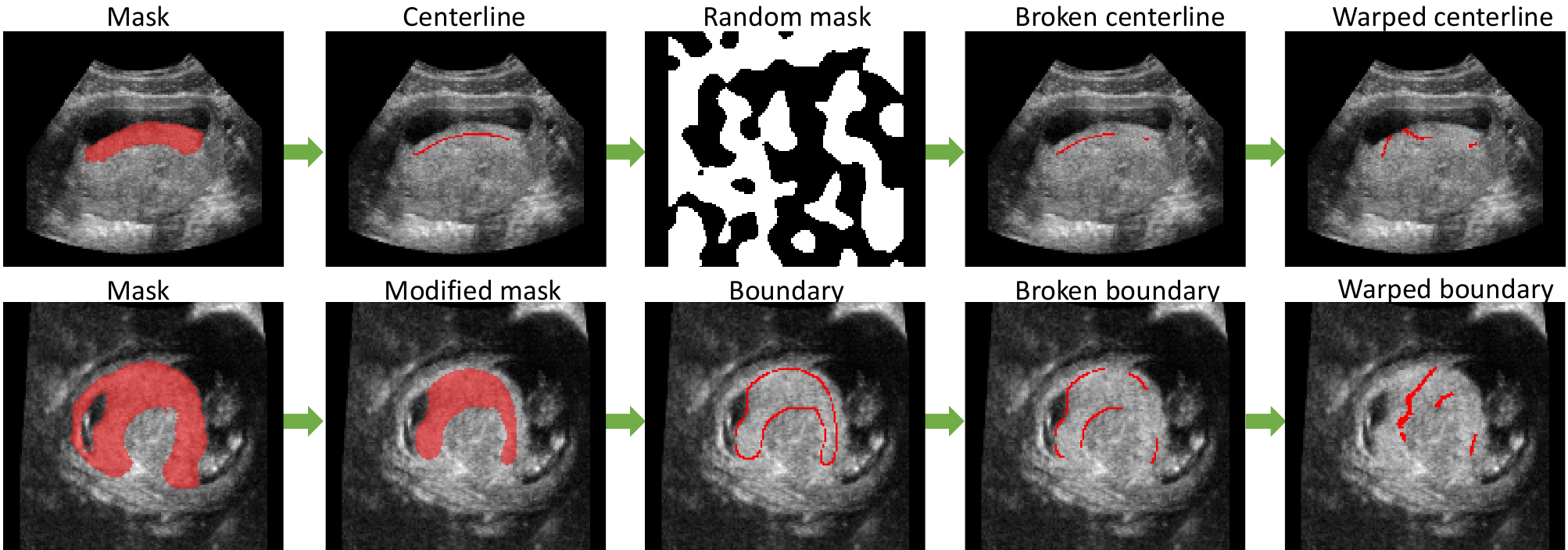}   
\caption{Top: Centerline scribble generated from a longitudinal slice. Bottom: Boundary scribble generated from a transverse view. The binary mask can be derived from either FP or FN masks to generate negative and positive scribbles, respectively. }

\label{scribbles}
\end{figure}

\noindent    \underline{\textbf{Centerline scribble}}: Scribble generation follows the ScribblePrompt \cite{wong2023scribbleprompt}. As depicted in Fig.~\ref{scribbles} top row, the first step involves identifying the centerline, which is the skeleton of a binary mask (e.g., the FN region for the current iteration). Next, a random binary mask is created to divide the centerline into separate, smaller parts. Lastly, the curvature and thickness of scribbles are modified using a random deformation field and Gaussian filtering.

\noindent     \underline{\textbf{Boundary scribble}}: As shown in Fig.~\ref{scribbles} bottom row, the modified binary mask is created from the original binary mask by applying a Gaussian blur and then thresholding with a random number between its minimum and maximum values. Next, the boundary of this modified mask is extracted. The methods used to generate broken and warped boundaries are same as in centerline generation.

These positive and negative scribbles are generated in a 2D manner from the slices of 3D binary FN and FP masks, respectively. We run the scribble generation algorithm (which may result in multiple broken scribbles as described above) once per FN/FP region, excluding small ($<100$ voxels) FN/FP regions.

We note that box prompts provide a strong prior but they are often inadequate to correct FP and FN regions. Conversely, points are highly flexible and widely used for their simplicity and efficiency. Finally, drawing scribbles is practical and often favored as an extension of point prompts, as they encode more information about user intentions. With these visual prompts, PRISM could handle different applications at various difficulty levels.

\subsection{Implementation details}
The training details and loss functions for PRISM follow \cite{li2024prism}.
We initially divided the data into training, validation, and testing with ratios of 0.6, 0.2, and 0.2, respectively. To assess the impact of varying training data ratios, we reduced the training ratio to 0.4 and 0.2, transferring the corresponding subjects (0.2 and 0.4) to the testing set. The subjects in the validation set remained unchanged.
The Dice, normalized surface Dice (NSD) with 1mm as tolerance \cite{kiser2021novel}, average symmetric surface distance (ASD), and 95-percent Hausdorff distance (HD95) are used as evaluation metrics, where distance are reported in mm. The study was conducted on an NVIDIA A6000.

\subsection{Compared methods}
We compare PRISM performance to other interactive state-of-the-art methods, including model adaptation methods \cite{li2023promise,gong20233dsam} and iterative methods for corrections \cite{wang2023sam}. The pretrained weights of SAM are utilized for model adaptation methods \cite{gong20233dsam,li2023promise}. The number of iterations is set to 11 to obtain the final segmentation using iterative methods for both training and inference \cite{wang2023sam}. We retrained the models with the hyperparameters using the official code for each, except for SAM, which was used only for inference. The prompt settings also follow their respective codes. The interactive segmentation methods show variability based on different prompts \cite{li2023assessing}, and all results were produced using fixed seeds to control for randomness. We also compare the performance with a state-of-the-art fully automated placenta segmentation method \cite{schwartz2022fully}.

\section{Results}

\begin{table*}[t]

\caption{$Dice / NSD$ results comparison. Bold indicates best performance.} 

\label{main_table}
\small
\begin{center}
    \begin{tabular}{c | c | l  | c |c |c }
    \hline
    \multicolumn{1}{l}{} & \multicolumn{1}{c}{Prompt} & \multicolumn{1}{l}{Methods} &   \multicolumn{1}{c}{Anterior} & \multicolumn{1}{c}{Posterior} & \multicolumn{1}{c}{Overall}\\
    \hline

    \multirow{1}{*}{} & - & Automated \cite{schwartz2022fully} &    90.46 / 80.72 & 89.42 / 75.75 & 90.03 / 78.65 \\



    \hline
    \multirow{7}{*}{\rotatebox{90}{Interactive}} & 10 points/slice & SAM \cite{Kirillov_2023_ICCV} & 46.44 / 14.29  & 43.32 / 14.78 &  45.14 / 14.49 \\ 



      & 1 point/volume & 3DSAM-adapter \cite{gong20233dsam}  & 84.57 / 70.34  & 85.97 / 68.58 & 85.15 / 69.61 \\



    & 1 point/volume &  ProMISe \cite{li2023promise}  &  85.55 / 71.57  & 83.91 / 67.15 & 84.87 / 69.73  \\
    

  & 1 point/volume &  SAM-Med3D \cite{wang2023sam} &  53.26 / 22.29  & 53.28 / 28.09 & 62.12 / 28.03  \\

  & 1 point/volume &  SAM-Med3D-turbo &  89.51 / 76.22  &  88.59 / 73.32 & 89.13 / 75.01  \\
    


  & 1 point/volume &  PRISM \cite{li2024prism} &  87.04 / 72.43   &  88.55 / 73.40 &  87.66 / 72.83  \\


     \cline{2-6}

   &  * &  PRISM &  \textbf{97.35 / 99.68}  &  \textbf{97.01 / 99.44}  & \textbf{97.15 / 99.54} \\
    \hline     
\multicolumn{6}{l}{
\begin{tabular}{@{}l@{}@{}} 
    * uses 1 point and 1 3D box per volume, as well as 2D warped centerline scribbles.
\end{tabular}
        }  
        
\end{tabular} 

\end{center}
\end{table*}

\begin{figure}[t]
\centering
\includegraphics[width=\linewidth]{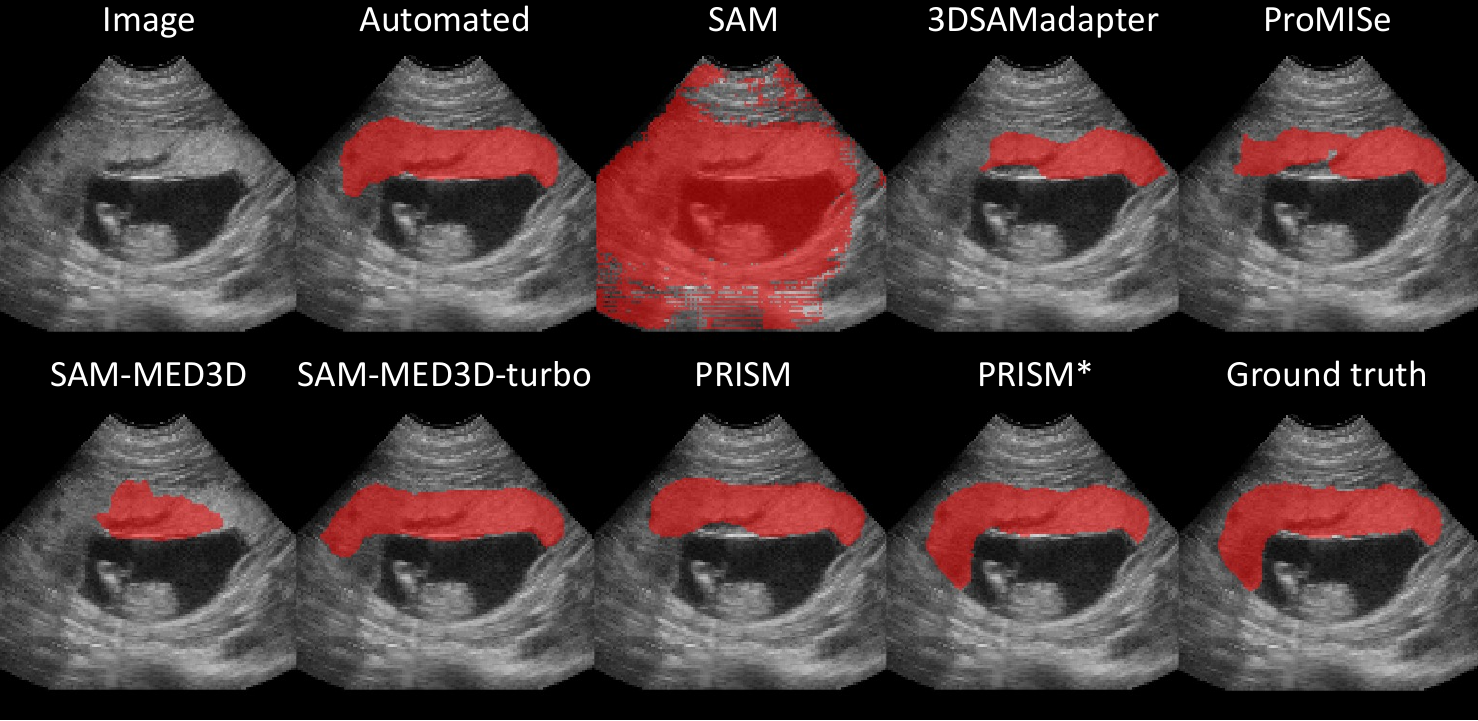}   
\caption{Qualitative results of compared methods for placenta segmentation.}
\label{qualitative_results}
\end{figure}

\noindent \underline{\textbf{Quantitative analysis.}}
The quantitative results are presented in Tab.~\ref{main_table}, where Dice and NSD are used as evaluation metrics. The automated method \cite{schwartz2022fully} can produce accurate placenta segmentation, with mean Dice scores slightly above 0.9. Interactive segmentation models tend to perform less effectively when using just one point prompt per volume, as all Dice scores for these methods are below 0.9. We note that, using a pretrained model (SAM-Med3D-turbo) significantly improves the accuracy of SAM-Med3D by leveraging the large dataset for pretraining. Among the methods using a single point prompt, PRISM outperforms these interactive models except the pretrained SAM-Med3D-turbo. PRISM achieves human-level performance when 2D scribble prompts are used, similar to the performance in non-ultrasound applications described in \cite{li2024prism}. The performance is slightly lower for posterior placentas, which is expected since these are harder to image and often contain shadowing artifacts and poorer contrast.

\noindent \underline{\textbf{Qualitative analysis.}}
The qualitative results are shown in Fig.~\ref{qualitative_results}. Major local defects are observed in all compared methods, except for PRISM with scribbles (denoted as PRISM* in Fig.~\ref{qualitative_results}), which produces the segmentation closest to the ground truth.

\begin{table*}[t]
\caption{Comparison of different scribble types for PRISM, presented as $Dice / NSD$. Bold  indicates best performance.  \ref{prompt_generation}.} 

\label{ablation_study}
\small
\begin{center}
    \begin{tabular}{ c | l  | c |c |c }
    
    \hline
  \multicolumn{1}{c}{View} &  \multicolumn{1}{c}{Methods} &   \multicolumn{1}{c}{Anterior} & \multicolumn{1}{c}{Posterior} & \multicolumn{1}{c}{Overall}\\
    \hline

   \multirow{4}{*}{Longitudinal} & Centerline &  \textbf{96.33 / 99.01}  & \textbf{96.87 / 99.58} & \textbf{96.56 / 99.25} \\
       
   & Warped centerline &  96.24 / 98.85  & 96.64 / 99.25 & 96.41 / 99.02 \\

  &  Boundary &  95.51 / 97.95  & 96.10 / 98.98 & 95.75 / 98.38 \\
  
  &  Warped boundary &  96.18 / 99.16  & 96.54 / 99.40 & 96.33 / 99.26 \\
\hline

     \multirow{4}{*}{Transverse} & Centerline &   \textbf{97.41 / 99.75}  &  \textbf{97.31 / 99.70}  & \textbf{97.55 / 99.82} \\

  & Warped centerline &  97.35 / 99.68  &  97.01 / 99.44  & 97.15 / 99.54 \\

   & Boundary &  96.05 / 98.85  & 96.48 / 99.21 & 96.22 / 99.00 \\

  & Warped boundary &  96.88 / 99.47  &  97.29 / 99.77  & 97.04 / 99.59 \\

    \hline

\end{tabular} 

\end{center}
\end{table*}

\begin{figure}[t]
\centering
\includegraphics[width=\linewidth]{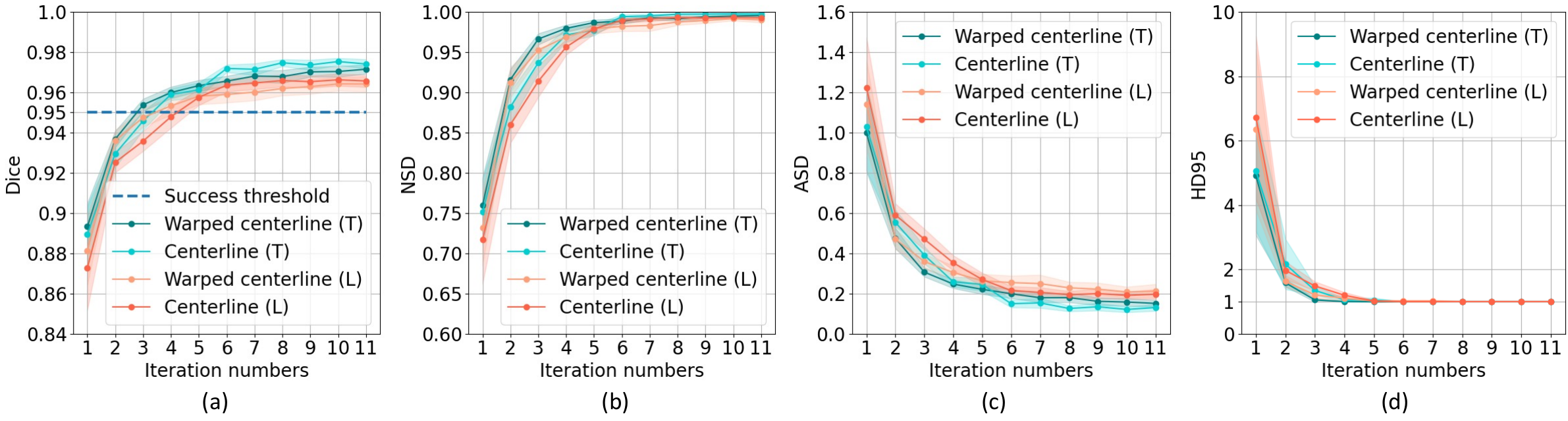}   
\caption{PRISM segmentation performance with different scribbles across iterations. (a) Dice, (b) NSD, (c) ASD, (d) HD95. The mean values (lines) and 95\% confidence intervals (shades) are shown. T and L denote transverse and longitudinal views, respectively.}
\label{iterative_results}
\end{figure}

\noindent \underline{\textbf{Comparison of scribble types.}} Next, we compare the different scribble settings for PRISM in Tab.~\ref{ablation_study}. Specifically, we explore performance variations by labeling scribbles on either longitudinal (Fig.~\ref{scribbles}(top)) or transverse (Fig.~\ref{scribbles}(bottom)) slices. In addition, we compare scribbles generated using the centerline, warped centerline, boundary, and warped boundary methods from Fig.~\ref{scribbles}.

We observe that generally, centerline scribbles are more effective than boundary scribbles, and generating them in the transverse slice achieves higher performance than in the longitudinal view.

We note that the centerline and boundary scribbles are perfect representations of the associated FN/FP region, whereas the warped variants introduce noise and randomness and are more representative of realistic human behavior. The warped variants also contain fewer points per scribble. Despite these two disadvantages, the performance of the various scribble types are comparable, which suggests that PRISM is effective even with fewer and imperfect prompts.
Even though it provides a slight performance boost, requiring an exact scribble is not a desirable approach in practice. Instead, the warped scribbles require less effort to generate and closely mimic human behavior.



\noindent \underline{\textbf{Iterative analysis.}} Next, we investigate the performance as a function of number of iterations (Fig.~\ref{iterative_results}). Regardless of the scribble type, the effectiveness of PRISM is significantly improved in the early iterations. The warped centerline reaches the 0.95 Dice score, which we consider indicative of a successful segmentation, by the third iteration.


    

    





\begin{table*}[t]
\caption{Sparse sampling results are presented as $Dice / NSD/ ASD / HD95$. We report both the performance on the whole 3D volume, as well as on just the slices with annotations. `Freq.' indicates prompt frequency, i.e., every 2 and 5 slices.} 

\label{propagation_table}
\small
\begin{center}
    \begin{tabular}{ c | c | c }
    
    \hline
    \multicolumn{1}{l}{$Freq.$} &   \multicolumn{1}{c}{Entire volume} &   \multicolumn{1}{c}{Annotated slices only} \\
    \hline

    1 & 97.15 / 99.8 / 0.15 / 1.00   &  -  \\ 
    
    2 & 96.71 / 99.2 / 0.19 / 1.00   &  97.09 / 99.9 / 0.02 / 0.00  \\ 

    5 &  95.28 / 96.1 / 0.32 / 1.23 &  96.64 / 99.6 / 0.02 / 0.13  \\

    \hline

\end{tabular} 

\end{center}
\end{table*}

\begin{figure}[b]
\centering
\includegraphics[width=\linewidth]{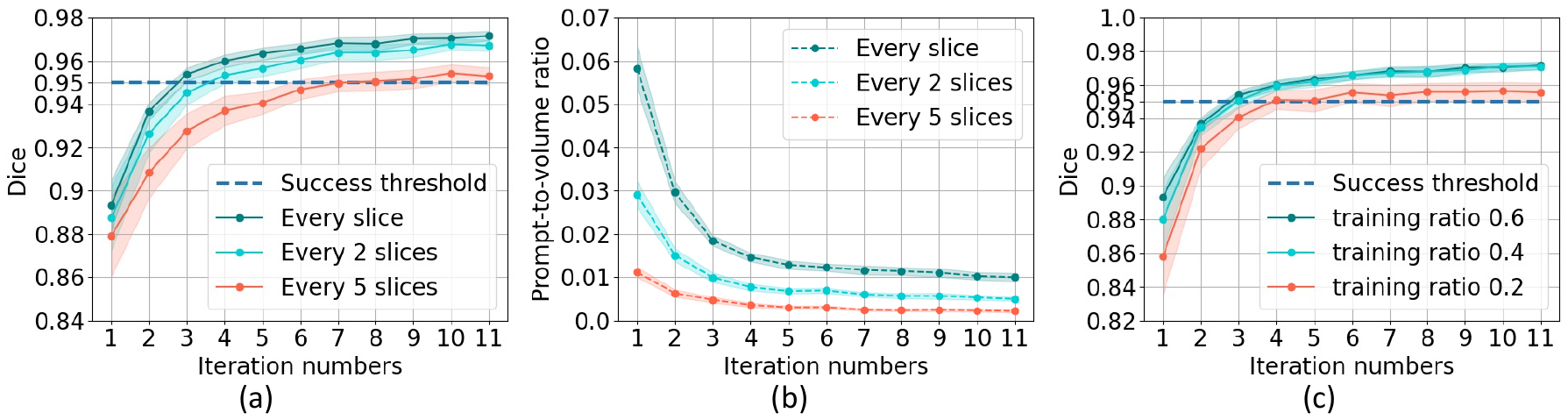}   
\caption{(a) Dice per iteration for sparse sampling. (b) Prompt-to-volume ratio  at each iteration for sparse sampling. The prompt-to-volume ratio is the ratio of the total number of scribble voxels to the number of placenta voxels in the ground truth. (c) Dice score on different training set size, without changing the test set.}
\label{sparse_and_split}
\end{figure}

\noindent \underline{\textbf{Sparse sampling.}} 
The results presented so far require scribbles to be generated for each 2D slice, which lowers the efficiency. Here, we investigate the performance of sparser scribbles by only generating them on a subset of slices.

Tab.~\ref{propagation_table} and Fig.~\ref{sparse_and_split}-a show the results with scribbles in every 2 and every 5 transverse slices. Compared to dense sampling, there is a slight decrease in performance, but it still reaches the 0.95 Dice bar. We also report the performance on only the slices with scribbles. Tab.~\ref{propagation_table} shows  that the results are slightly better in these annotated slices compared to the rest of the volume. Exploring strategies for either propagating the sparse prompts prior to the segmentation, or propagating the segmentation from the annotated slices to the other slices remains as future work.  Fig.~\ref{sparse_and_split}-b shows the ratio of scribble voxels to placenta volume for each setting. We observe that the required prompt voxels quickly goes down after the first few iterations. Overall, we conclude that PRISM can produce robust segmentations with even sparsely labeled scribbles.


\noindent \underline{\textbf{Smaller training dataset.}} Finally, we explore the robustness of the model to smaller training datasets. The training data ratio in the experiments presented so far was set at 0.6, and here we reduce it to 0.4 and 0.2. The test subjects remain the same for a fair comparison. The results are presented in Fig.~\ref{sparse_and_split}-c. Despite having fewer training images, PRISM can maintain similar performance and reach 0.95 Dice in just a few iterations, showing its effectiveness and efficiency. 

\section{Conclusion}
In this study, we present a comprehensive evaluation of deep learning-based 3D interactive segmentation models for placenta segmentation, specifically focusing on the effectiveness and efficiency of PRISM in terms of segmentation quality and the effort required to generate prompts. With scribbles and a few iterative corrections, PRISM can produce segmentations that reach human-level performance. The results show both effectiveness and efficiency in responding to scribbles. Future work will involve  a pretrained model to further improve efficiency.

\noindent
\textbf{\textit{Acknowledgments.}}
This work was supported, in part, by NIH grants R01-HD-109739, R01-HL-156034, U01-HD-087180, and R03-HD-069742-02, as well as by the Penn Presbyterian George L. and Emily McMichael Harrison Fund for Research in Obstetrics and Gynecology.


%
%
%
\bibliographystyle{splncs04}
\bibliography{mybibliography}

\end{document}